\documentclass[letterpaper]{article} 
\usepackage{aaai2026}  
\usepackage{times}  
\usepackage{helvet}  
\usepackage{courier}  
\usepackage[hyphens]{url}  
\usepackage{graphicx} 
\usepackage{natbib}  
\usepackage{caption} 
\usepackage{algorithm}
\usepackage{algorithmic}
\usepackage{amsmath}

\usepackage{newfloat}
\usepackage{multirow} 
\usepackage{adjustbox}
\usepackage{listings}
\usepackage{bm} 
\DeclareCaptionStyle{ruled}{labelfont=normalfont,labelsep=colon,strut=off} 
\floatstyle{ruled}
\newfloat{listing}{tb}{lst}{}
\floatname{listing}{Listing}
\usepackage{booktabs}
\usepackage{xcolor}         
\usepackage{tabularx}
\usepackage{makecell} 
\usepackage{array} 
\newcommand{\cmark}{\includegraphics[height=1.5ex]{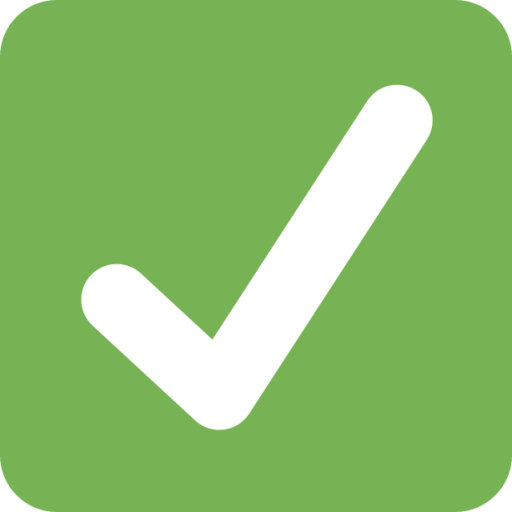}}
\newcommand{\xmark}{\includegraphics[height=1.5ex]{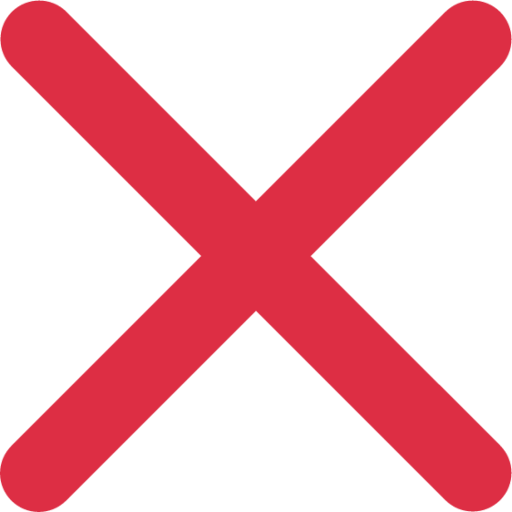}}
\usepackage{amsfonts}

\usepackage{scrextend}
\title{Aligning LLMs on a Budget: \\
Inference-Time Alignment with Heuristic Reward Models}
\author {
    Mason Nakamura\equalcontrib,~
    Saaduddin Mahmud\equalcontrib,~
    Kyle H. Wray, ~
    Hamed Zamani, ~
    Shlomo Zilberstein
}
\affiliations {
    Manning College of Information and Computer Sciences\\
    University of Massachusetts Amherst, USA 
}

\begin{document}
\nocopyright
\maketitle
\begin{abstract}
    Aligning LLMs with user preferences is crucial for real-world use but often requires costly fine-tuning or expensive inference, forcing trade-offs between alignment quality and computational cost. Existing inference-time methods typically ignore this balance, focusing solely on the optimized policy's performance. We propose \textsc{HIA} (Heuristic-Guided Inference-time Alignment), a tuning-free, black-box-compatible approach that uses a lightweight prompt optimizer, heuristic reward models, and two-stage filtering to reduce inference calls while preserving alignment quality. On real-world prompt datasets, \textsc{HelpSteer} and \textsc{ComPRed}, \textsc{HIA} outperforms best-of-N sampling, beam search, and greedy search baselines in multi-objective, goal-conditioned tasks under the same inference budget. We also find that \textsc{HIA} is effective under low-inference budgets with as little as one or two response queries, offering a practical solution for scalable, personalized LLM deployment.
\end{abstract}

\section{Introduction}
Value alignment is the process of encoding values, objectives, and goals into a model so that its behavior reflects that of the users' beliefs, preferences, and expectations. Deploying Large Language Models (LLMs) at scale requires not only aligning model behavior with user values, but controlling the per-query cost of that alignment. Current techniques such as Reinforcement Learning from Human Feedback (RLHF) \cite{ouyang2022training, ziegler2019fine, stiennon2020learning} and Direct Preference Optimization (DPO) \citep{rafailov2023direct} have been used to mitigate unintended risks and achieve stronger user alignment by fine-tuning model weights, yet this optimization is infeasible for models without weight access (i.e., black-box models). This makes them incompatible with inference-time alignment. Moreover, these methods still optimize a single scalar reward, which has been shown to be theoretically inadequate \cite{vamplew2022scalar, chakraborty2024maxmin} for aligning systems with users who have diverse or conflicting preferences.

\begin{figure*}[!t]
    \centering
    \includegraphics[width=6.6in]{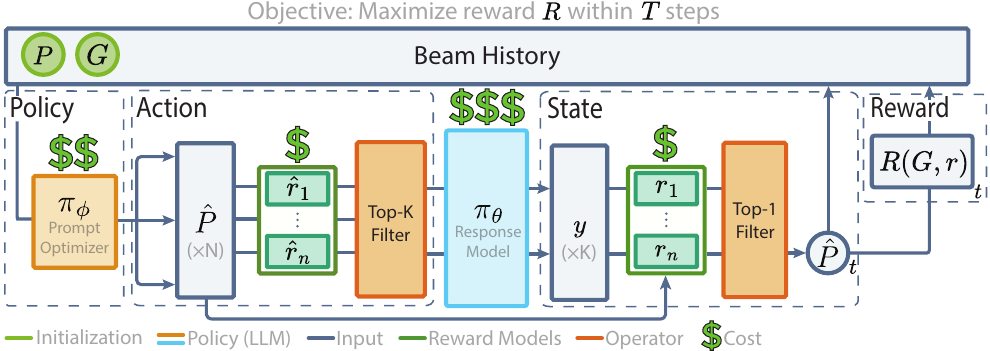}
    \caption{\textbf{\textsc{HIA} Framework Applied to Beam Search.} Upon initialization, a prompt $P$ and goal reward vector $\bm{G}$ is provided. The policy $\pi_\phi$ that acts as a prompt optimizer performs $N$ prompt modifications $\hat{P}$ of the original prompt $P$, all of which are then scored using the heuristic reward models $\{\bm{\hat{r}}\}_{i=1}^{n}$ for $n$ objectives. The top-$K$ modified prompts are filtered using the reward function $R(\bm{G}, \bm{\hat{r}})$ where $\bm{\hat{r}}$ is the heuristic reward vector for the modified prompt. The top-$K$ modified prompts are given as input to the response model, generating $K$ responses. Next, each response is scored using the reference reward models $\{r\}_{i=1}^{n}$, and the modified prompt $\hat{P}$ with the highest reward from the reward function $R$ being used for the next beam search step.}
    \label{fig:HIA-framework}
\end{figure*}

To address these issues, recent work has explored inference-time alignment methodologies that employ search and sampling procedures to align responses at inference, making them compatible with black-box models. This is especially important for state-of-the-art LLMs \cite{comanici2025gemini25pushingfrontier, openai2024gpt4technicalreport} that restrict model-weight access to maintain competitive advantage \cite{habibi2025open, schrepel2024competition}. While these models excel on natural language tasks, coding \cite{jiang2024survey}, and benchmarks (e.g., Chatbot Arena \cite{chiang2024chatbot}, GSM8K \cite{cobbe2021gsm8k}, etc.), they do not allow unrestricted fine-tuning. Prompt optimization \cite{cheng2023black, singla2024dynamic} has emerged as an effective proxy, modifying the prompt at a token level in an interpretable way to improve alignment without modifying model weights. This allows inference-time scaling for alignment but comes at the cost of many expensive inference samples.

Beyond this cost issue, scalar reward functions introduce their own limitations; they often reflect the average annotator's preference \cite{chakraborty2024maxmin}, potentially masking important minority viewpoints, reducing personalization, and resulting in suboptimal misalignment outcomes. To address this, recent work introduces multi-objective alignment approaches \cite{rame2023rewarded,chakraborty2024maxmin,zhou2023beyond,yang2024rewards,wang2024arithmetic}, reframing alignment as a multi-objective problem using multiple reward models. By explicitly modeling multiple objectives (e.g., verbosity, coherence, political ideologies), these approaches enable more nuanced and personalized alignment \cite{salemi2023lamp} for users with diverse backgrounds. Despite this progress, most existing multi-objective approaches favor static scalarizations (e.g., sums of weighted rewards) or Pareto-based maximization, which limits personalization of specific user goals.

Although recent work has made progress in addressing the challenges of black-box alignment and the limitations of scalar reward optimization, key issues remain. In particular, existing approaches often incur high inference costs during inference-time alignment and fail to extend multi-objective alignment frameworks to support goal-conditioned optimization at inference time for personalization. To address these gaps, we introduce \textsc{HIA} (Heuristic-Guided Inference-time Alignment), a tuning-free, goal-conditioned alignment framework that works with any prompt-optimization-based search or sampling strategy. \textsc{HIA} leverages lightweight heuristic reward models to enable efficient inference-time alignment, achieving better sample efficiency compared to their non-heuristic enabled counterparts under equivalent compute budgets. Moreover, by conditioning on user-specific goals, \textsc{HIA} supports fine-grained personalization across diverse user preferences.
\vspace{-1mm}
In real-world applications, practitioners and engineers deploying inference-time alignment methods must balance the trade-off between alignment quality and inference cost, costs attributed to financial charges, energy usage, or carbon emissions. This challenge is especially pronounced when working with commercial, state-of-the-art LLMs \cite{comanici2025gemini25pushingfrontier, openai2024gpt4technicalreport} that are accessible only through costly APIs and cannot be fine-tuned nonrestrictively. In such settings, each query to the response model incurs a significant cost, making it essential to minimize expensive model calls while maintaining high-quality alignment. In contrast, prompt optimizers can often run locally or with cheap servicing given their size while we treat the cost of scoring models as negligible, providing a response-heavy inference-cost trade-off between the prompt optimizer and the response model (i.e., calls to the large LLM dominates majority of costs). We show \textsc{HIA} is specifically designed for this response-heavy inference-cost trade-off under very constrained inference budgets by filtering out poor prompt modification candidates prior to expensive response model inference.

\begin{table*}[t]
\vspace{-1mm}
  \centering
  \begin{tabular}{@{}lcccccc@{}}
    \toprule
    Method & Methodology & Search-Agnostic & Tuning-Free & Multi-Objective & Tunes \_\_\_\\
    \midrule
    RLHF \cite{ouyang2022training}              & PPO & \xmark & \xmark & \xmark & Policy \\
    RiC \cite{yang2024rewards}                  & SFT & \xmark & \xmark & \cmark & Policy \\
    MetaAligner \cite{yang2024metaaligner}      & SFT & \xmark & \xmark & \cmark & Response \\
    DRPO \cite{singla2024dynamic}               & Beam Search & \xmark & \cmark & \cmark & Prompt\\
    Speculative Rejection \cite{sun2024fast}    & Filtering & \xmark & \cmark & \xmark & N/A\\
    \textsc{HIA} (ours)                         & Filtering & \cmark  & \cmark & \cmark & Prompt\\
    \bottomrule
  \end{tabular}
    \caption{\textbf{Related Methods}. \textsc{HIA} is agnostic to the sampling or search procedure employed for inference-time alignment, does not require fine-tuning during the post-training phase, and is applicable to goal-conditioned multi-objective domains.}
    \label{tab:method-comparison}
\end{table*}

\textbf{Main Contributions.} We propose \textsc{HIA}, an inference-time alignment framework that (i) improves goal-conditioned multi-objective alignment for explicit personalization, (ii) is built on a search and sampling-agnostic framework that requires no fine-tuning, allowing application to black-box models, and (iii) provides improved inference-time alignment under very low-inference budgets (i.e., one or two queries) compared to existing inference-time methodologies.

\section{Related Work}
\textbf{Learning From Human-Feedback.} The process of fine-tuning a model to elicit responses that align with human preferences, also known as Reinforcement Learning From Human-Feedback (RLHF) \cite{ouyang2022training, ziegler2019fine, stiennon2020learning}, has been widely adopted to steer foundation models. After a model is supervised fine-tuned (SFT) on a down-stream task (e.g., summarization, classification), a reward model is learned from human preferences, and the SFT model is fine-tuned using reinforcement learning (RL) on the learned reward model which typically involves using Proximal-Policy Optimization \cite{schulman2017proximal} to penalize large changes in the policy distribution. 

\textbf{Multi-Objective Alignment from Human Feedback.} RLHF optimizes a policy using a single, monolithic reward model trained on human preference data. In multi-objective RL from human-feedback (MORLHF), multiple reward models are learned which can represent textual attributes (e.g., complexity, verbosity) or complex group preferences (e.g., political ideology). Using these reward models, a policy can be trained using policy-weight interpolation \cite{rame2023rewarded} to combine multiple learned policies, scheduled reward interpolation policy-gradient \cite{chakraborty2024maxmin} for fair alignment, multi-objective direct preference optimization \cite{zhou2023beyond} to avoid RL and reward modeling, or multi-conditional reward SFT \cite{yang2024rewards} for inference-time alignment while training a single policy without RL. Rewards-in-Context (RiC) \cite{yang2024rewards} is a multi-objective alignment approach that uses SFT and rejection sampling for inference-time alignment. It directly fine-tunes the response model, whereas we treat the response model as non-fine-tunable, posing a more realistic problem formulation for aligning black-box LLMs. In our work, we take a goal-conditioned evaluation approach of multi-objective rewards adopted from goal-conditioned RL \cite{kobanda2025offline}, whereas prior MORLHF approaches used static scalarizations or Pareto-based maximization which is not as effective for personalization that requires goals.

\textbf{Inference-Time Alignment.} Inference-time alignment is the process of aligning a model's outputs during inference time which has gained traction for its scaling potential and simplicity. Most inference-time alignment methods are categorized as tuning-free, meaning they do not require fine-tuning the target model's policy to elicit alignment. In inference-time alignment, there tends to exist two general categorizations during inference-time, response \cite{yang2024metaaligner, chen2024pad} and prompt \cite{cheng2023black, brown2020language, lester2021power, singla2024dynamic, pryzant2023automatic} optimization. In response optimization, the alignment algorithm modifies the model outputs or logits to maximize a preference reward function or a set of multi-objective reward functions. For example, MetaAligner \cite{yang2024metaaligner} modifies the output tokens of a black-box model to facilitate better output alignment using a correction module trained with SFT using a cross-entropy loss on weak-to-strong preference pairs. In contrast, prompt optimization methods tune the prompt in token \cite{lester2021power} or embedding \cite{brown2020language} space, modifying the outputs while remaining within the model's token output distribution. To take advantage of inference-time scaling, many prompt optimization works have explored the use of heuristic search algorithms with textual gradients \cite{wang2023promptagent} to effectively leverage inference such as Monte-Carlo Tree Search \cite{wang2023promptagent}, beam search \cite{pryzant2023automatic, singla2024dynamic}, and evolutionary algorithms \cite{yang2023large}. Most relevant to our work is \textsc{Speculative Rejection} \cite{sun2024fast} that reduces computation costs at decoding-time for BoN sampling using a reward model that identifies continuations that are poor and unlikely to be high-scoring. They focus on decoding-time efficiency and do not edit prompts. In contrast, our \textsc{HIA} operates before decoding: a prompt optimizer proposes many edits, heuristic reward models score them, and only the top-$K$ prompts are decoded with the response model.

\section{Background}
    \textbf{Markov Decision Process.} We consider a Finite-Horizon Markov-Decision Process (MDP)\footnote{We omit the discounted variation of the MDP since we only perform policy evaluation} defined by the tuple $(S, A, \mathcal{T}, \mathcal{G}, R)$ \cite{puterman2014markov}. Let $T$ denote the finite token space and $*$ as the Kleene star, then $S= \mathbb{R}_G^n \times \mathbb{R}_r^n \times T_P^* \times T_y^*$ is the state space containing the goal rewards vector $\bm{G}\in \mathbb{R}^n$, the rewards vector $\bm{r} \in \mathbb{R}^n$,
    the tokens for the prompt $P\in T^*$, and the response $y\in T^*$. The rewards vector $\bm{r}$ is generated by a set of reward models $\{r_i\}_{i=1}^n$ such that $r_i: T_P^* \times T_y^*\rightarrow \mathbb{R}$ is a mapping from a prompt and response pair to a real-valued reward.  Now, we define $A= T_{\hat{P}}^*$ as the action space containing the tokens of the modified prompt $\hat{P}\in T^*$.
    
    A stochastic policy $\pi: S \rightarrow A$, is a mapping from states to actions. In \textsc{HIA}, we hold three stochastic policies, $\pi_\phi$ as the prompt optimizer, $\pi_F$ as the filtering policy, and $\pi_\theta$ as the response model. To consolidate notation, we combine $\pi_\phi$ and $\pi_F$ into a joint action policy as $\pi_\textsc{jnt}(a\mid s) = \sum_{a'\in A}\pi_F(a\mid a', s)\pi_\phi(a'\mid s)$. We define the history \( h_t \) as a sequence of states and actions up to time \( t \), $(s_0, a_0 \dots, s_t)$. Next, $\mathcal{T}(s_{t+1}\mid s_{t}, a_{t})=\pi_\theta(y_{t+1}\mid a_{t}) \cdot p(r_{t+1} \mid y_{t+1})$ are the transition dynamics where $\pi_\theta$ gives the probability of selecting a response $y_{t+1}$ given a modified prompt $a_t$, and $p$ is the probability of obtaining the reward vector $\bm{r}_{t+1}$ given $a_{t+1}$ and $y_{t+1}$ from the reward models $\{r_i\}_{i=1}^n$.
    
    We denote $R: S \times A \times S\rightarrow \mathbb{R}$ as the reward function that signals how close the reward vector $\bm{r}_t$ in the state $s_t$ is to the goal reward vector $\bm{G}$ for all states $s$. More specifically, under the finite-horizon definition, the reward is $0$ if the state has reached the goal (i.e., $\bm{r} = \bm{G}$ where $=$ is element-wise equivalence) or if the time-step exceeds the horizon $T\in \mathbb{Z}^+$:
    \[
        R(s_t, a_t, s_{t+1}) =   \begin{cases}
                          R(s_t, a_t, s_{t+1}) & \text{if } \bm{r} \neq \bm{G} \text{ and } t\leq T,\\    
                          0 & \text{else. }
                        \end{cases}
    \]
     We define a trajectory, $\tau$, as a finite sequence of state, action, and rewards $\tau=(s_0, a_0, \bm{r}_0, \cdots)$. A policy $\pi$ induces a value function $V^\pi: S \rightarrow \mathbb R$ which gives the expected cumulative return, $V^\pi(s)=\mathbb{E}_\pi[\sum_{t=0}^\infty R(s_t)\mid s_0=s]$, when starting from state $s$ and induced by $\pi$. An optimal policy $\pi^*$ maximizes the expected cumulative return $V^*(s)$ for any state $s$. Later, we show \textsc{HIA} improves the action selection process by obtaining larger value when employing a heuristic filtering policy $\pi_F$ compared to a baseline filtering policy.

    \textbf{RLHF Pipeline.}
    To consider how \textsc{HIA} fits into the current value alignment literature, we briefly explain the RLHF pipeline.
    In the RLHF pipeline, the first stage is \textit{pre-training} where an LLM is trained on large-scale unlabeled data using a cross-entropy loss for next-token prediction. After pretraining, the learned base model is fine-tuned for a down-stream task during the \textit{supervised fine-tuning} (SFT) phase (e.g., classification, summarization), outputing an SFT model. Next, is the \textit{reward modeling} phase where a reward model is learned on human preference data to evaluate good and bad responses from the SFT model. Finally, the \textit{post-training} phase applies a policy gradient algorithm, typically PPO \cite{schulman2017proximal}, to optimize the SFT model using the learned reward model as a reward function.

    \textbf{Reward Modeling.} 
    In preference learning, the objective is to approximate the ground-truth preference model, $p^*:\mathcal{X} \times \mathcal Y \longrightarrow \mathbb{R}$, where $\mathcal{X}$ is the set of prompts and $\mathcal{Y}$ is the set of responses. In practice, learning $p^*$ requires fitting a Bradley-Terry-Luce (BTL) model \citep{Bradley1952RankAO} with maximum likelihood estimation (MLE) over a preference dataset $\mathcal{D}$. Here, $\mathcal{D}=\{\mathbf{y^{(i)}_w}, \mathbf{y^{(i)}_l}, \mathbf{x^{(i)}}\}^{n}_{i=1}$ which contains the preferred responses $\mathbf{y_w}$, the rejected responses $\mathbf{y_l}$, and the input $\mathbf{x}$. Then, the preference function modeled using a BTL model, with respect to a parameterized reward model $r_\phi$, can be defined as
    \[
        p(\mathbf{y_1} \succ \mathbf{y_2} \mid \mathbf{x}) = \frac{\exp(r_{\phi}(\mathbf{y_1}, \mathbf{x}))}{\exp(r_{\phi}(\mathbf{y_1}, \mathbf{x})) + \exp(r_{\phi}(\mathbf{y_2}, \mathbf{x}))} \label{eq:preference_model}
    \]
    where the reward model is the mapping $r_\phi: \mathcal{X} \times \mathcal{Y} \rightarrow \mathbb{R}$.
     
    \textbf{Heuristic Reward Modeling.}
    A typical reward model uses both the response $y\in \mathcal{Y}$ and the prompt $x\in \mathcal{X}$ to infer a reward. However, this \textit{reference reward model} can be approximated by removing or replacing the response feature, which is preferable in practice when considering inference costs but this may degrade reward model accuracy. More precisely, we define this \textit{heuristic reward model} as $\hat{r}_\phi: \mathcal{X} \times \hat{\mathcal{Y}} \rightarrow \mathbb{R}$ where the set $\hat{\mathcal{Y}}$ is the empty set or the set of proxy features. In this situation, a proxy feature could be the model ID of the response model, offering a feature of the output distribution that is less computationally expensive compared to the inference needed to produce a response.
    
   \textbf{Supervised Fine-Tuning.}
    Supervised fine-tuning (SFT) is a technique to adapt a pre-trained language model to a specific downstream task (e.g., summarization, reward modeling) or domain using labeled data and a cross-entropy loss.

 In our framework, we apply SFT to train heuristic reward models; however, this process can be done in parallel to training reward models during the reward modeling phase, leaving minimal overhead.

\section{The \textsc{HIA} Framework}
We now present a precise description of our framework as seen in Figure \ref{fig:HIA-framework}. The \textsc{HIA} framework can be defined by the following tuple $(R, \mathcal{R}, \hat{\mathcal{R}}, \pi_\phi, \pi_\theta, \pi_F, K, N)$:
\begin{itemize}
    \item $R$: The reward function (e.g., goal completion, Euclidean distance, max-error) in the MDP used to evaluate policy performance using the current state's goal reward vector and inferred reward vector.
    \item $\mathcal{R}$: A set of $n$ reward models $\mathcal{R}=\{r_i\}_{i=1}^n$, each representing an objective (e.g., verbosity, coherence, political ideology, etc.) where $r_i: T_P^* \times T_y^*\rightarrow \mathbb{R}$ is a mapping from a prompt and response pair to a scalar reward.
    \item $\hat{\mathcal{R}}$: A set of $n$ heuristic reward models $\hat{\mathcal{R}}=\{\hat{r}_i\}_{i=1}^n$,  where $\hat{r}_i: T_P^* \times T_\textsc{ID}^* \rightarrow \mathbb{R}$ is a mapping from a prompt and the model ID of the response model to a scalar reward.
    \item $\pi_\phi$: The prompt optimizer policy that adjusts the initial prompt according to the relevant objectives and their descriptions, along with the desired goal reward vector $\bm{G}$. We assume the prompt optimizer policy $\pi_\phi$ is provided, which can be derived from an out-of-the-box LLM\footnote{Preferably instruction fine-tuned} or an SFT model tuned for prompt optimization.
    \item $\pi_\theta$: The black-box response model that outputs a response $y$ according to a prompt $P$.
    \item $\pi_\textsc{F}$: A filtering policy that selects a top-$K$ from $N$ candidates.
    \item $K$: The number queries to the response model $\pi_\theta$ and reward models $\mathcal{R}$ in a top-$K$ best-of-$N$ setup.
    \item $N$: The number of prompt samples used to query the heuristic reward models $\hat{\mathcal{R}}$ where $K \leq N$.
\end{itemize}

To properly evaluate the performance of goal-conditioned alignment, especially in the context of multi-objective alignment, we experiment with various types of reward functions within our value function to evaluate the filtering policy $\pi_\textsc{jnt}$ under different sampling and search procedures. For our specific policy evaluation use, we define the value function as
\[
    V^{\pi_\textsc{jnt}}(s_t) = \max_{a_t\in A} Q^{\pi_\textsc{jnt}}(s_t, a_t)
\]
where $Q^{\pi_\textsc{jnt}}(s_t, a_t)$ is the action-value function,
\[
    Q^{\pi_\textsc{jnt}}(s_t, a_t) = \mathbb{E}_{y\sim \pi_\theta(\cdot\mid a_t)}\left[R\left(G, \left[r_i(y,a_t)\right]_{i=1}^n\right)\right].
\]

We now define the reward functions, $R$, used for our policy evaluation of $V^{\pi_\textsc{jnt}}(\cdot)$.

\textbf{Euclidean (L2) Distance.}
We define the Euclidean distance metric as the Euclidean distance between a goal vector $\bm{G}$ and a reward vector $\bm{r}$, and we apply a negative coefficient to assert maximization. 
\[
    R_{L2}(\bm{G},\bm{r})=
    -\lVert \bm{r}-\bm{G}\rVert_2.
\]

This metric, commonly used in heuristic search \cite{hart1968formal} and goal-conditioned reinforcement learning \cite{kobanda2025offline}, offers insight into how close a state is to the goal.

\textbf{Goal Completion.} A goal completion occurs when a goal vector $\bm{G}$ and a reward vector $\bm{r}$ are element-wise equivalent,
\[
    R_{\text{completion}}(\bm{G},\bm{r})=
    \mathbb{I}\!\left[\lVert \bm{r}-\bm{G}\rVert_\infty = 0\right].
\]
 This metric is commonly used in goal-conditioned deep reinforcement learning as an indicator for determining goal achievement of an agent \cite{qiu2023instructing}. The goal completion indicator emphasizes exact completion to the goal, which is an interpretable measurement of success.

\textbf{Maximum Error.} The max-error or worst-case alignment reward gives the worst-case value difference between all elements in the goal vector $\bm{G}$ and the reward vector $\bm{r}$. Here, we apply a negative coefficient to assert maximization,
\[
    R_{\text{max-error}}(\bm{G},\bm{r})=
    -\lVert \bm{r}-\bm{G}\rVert_\infty.
\]

A metric commonly employed in fairness and multi-objective optimization such as minimizing the maximum error \cite{chakraborty2024maxmin} and safe reinforcement learning \cite{achiam2017constrained} for conditioning on the worst-case reward scenario.

The full algorithm procedure of \textsc{HIA} is summarized in Figure \ref{fig:HIA-framework}. Together, the joint prompt optimization and heuristic-filtering policies of \textsc{HIA} aim to reduce costly inference calls to the response model without compromising performance. In the following Experiments section we outline the evaluation protocol, describe the \textsc{HelpSteer} and \textsc{Political} reward models, specify hyper-parameters that equalize compute budgets across methods, and report results on goal-completion, max-error, and L2 distance metrics.

\section{Experiments}
In this section, we evaluate the efficacy of \textsc{HIA} against three inference-time, tuning-free alignment baselines on two real-world reward model domains and show (1) \textsc{HIA} improves alignment performance under equal inference budgets compared to baselines on three reward functions for goals containing up to three objectives, (2) \textsc{HIA} is highly sample-efficient at low-inference budgets, and (3) \textsc{HIA} improves alignment performance on multiple goal-conditioned objectives, showing personalization improvement. We specifically show that \textsc{HIA} improves alignment performance for a response model inference budget of $K=1$, one query, while achieving as much as a 29\% improvement on goal completion reward for BoN sampling. 
\subsection{Reward Models}
To acquire accurate reward models of objectives that represent real human values, we train all reward models on human-labeled data that were annotated via a human-labeler such as HelpSteer or on pairs of naturally occurring human-generated text from Reddit.

\textbf{HelpSteer Objectives.}
Objectives that mostly hold no dependence on the subject of the text such as verbosity, coherence, complexity, helpfulness, and correctness serve as useful features for personalization on formatting. 
The HelpSteer datasets, HelpSteer 
\cite{wang2023helpsteer, dong2023steerlm} and HelpSteer2 \cite{ wang2024helpsteer2preferencecomplementingratingspreferences, wang2024helpsteer2}, comprise of 55k total samples, each containing a prompt, response, and 5 human-labeled attributes of the response each ranging between 0 and 4. The attributes we use in this paper are complexity, verbosity, and coherence. We trained a reference reward model for each attribute using supervised fine-tuning on a 500M parameter ModernBERT model \cite{modernbert} with a cross-entropy loss. Although helpfulness and correctness were included in the dataset, we omit these attributes because the learned reference reward models tend to be noisy and less grounded to objective features of the response.

\textbf{Political Ideologies.}
For political ideologies, we use real-world Reddit data from the \textsc{ComPRed} 
dataset~\cite{kumar2024compocommunitypreferenceslanguage}, taking three political ideologies from their respective subreddits (see Appendix for details). Using pairs of text from different subreddits as the dataset, we supervise fine-tune a 500M parameter ModernBERT model \cite{modernbert} as a reference reward model for each ideology by fitting a BTL Model on the training dataset using MLE. We then normalize the reward distribution from a normal distribution to a uniform distribution since the reward distribution was skewed due to preference learning, as opposed to the HelpSteer dataset that was trained using SFT.

\textbf{Heuristic Reward Models.}
Each reference reward model (RRM) has an associated heuristic reward model (HRM) where the HRMs were fine-tuned on 500M parameter ModernBERT models in similar fashion to the RRMs. However, we replace the response feature $\mathbf{y}$ used in the reward models as seen in Equation \ref{eq:preference_model} with a static response model ID, allowing the HRM to adapt to different response models while also reducing heuristic reward inference.

\begin{table*}[t]
  \centering
  \setlength\tabcolsep{5pt}
  \renewcommand{\arraystretch}{1.0}
  \begin{adjustbox}{width=.95\textwidth,center}
  \begin{tabular}{@{}clccccccc@{}}
    \toprule
    & & \multicolumn{3}{c}{\textsc{HelpSteer}} & & \multicolumn{3}{c}{\textsc{Political}}\\[-2pt]
    \cmidrule(lr){3-5}\cmidrule(lr){7-9}
    Reward & Method & 1 Obj & 2 Objs & 3 Objs & & 1 Obj & 2 Objs & 3 Objs\\
    \midrule
    \multirow{7}{*}{\centering Goal Comp.\,(\%)} 
      & \textsc{MF-BoN+Random} & $21.67 \pm 4.78$ & $4.00 \pm 0.82$ & $0.33 \pm 0.47$
      && $17.67 \pm 4.11$ & $3.33 \pm 1.25$ & $0.00 \pm 0.00$ \\
    & \textsc{BoN+Random}
      & $24.00 \pm 4.32$ & $4.50 \pm 1.50$ & $0.67 \pm 0.47$
      && $17.0 \pm 2.94$ & $3.67 \pm 0.94$ & $0.00 \pm 0.00$ \\
    & \textsc{BS+Random}
      & $25.33 \pm 3.30$ & $4.33 \pm 0.94$ & $1.33 \pm 0.94$
      && \bm{$25.00 \pm 5.89$} & $4.67 \pm 2.62$ & $0.33 \pm 0.47$ \\
    & \textsc{GS+Random}
      & $26.67 \pm 5.79$ & $7.67 \pm 1.70$ & $1.50 \pm 0.50$
      && $24.33 \pm 6.18$ & $6.00 \pm 2.83$ & $0.33 \pm 0.47$ \\
      [-2pt] \cmidrule(lr){2-9}
    & \textsc{BoN+H} (ours)
      & \bm{$31.00 \pm 3.00$} & $6.33 \pm 2.87$ & \bm{$2.33 \pm 1.25$}
      && $20.67 \pm 3.30$ & $3.33 \pm 0.94$ & $0.00 \pm 0.00$ \\
    & \textsc{BS+H} (ours)
      & $30.00 \pm 2.16$ & $6.33 \pm 1.25$ & $1.00 \pm 0.82$
      && $21.00 \pm 1.00$ & $3.33 \pm 0.47$ & \bm{$1.00 \pm 0.00$} \\
    & \textsc{GS+H} (ours)
      & $30.00 \pm 4.55$ & \bm{$8.33 \pm 1.70$} & $1.67 \pm 0.94$
      && $21.00 \pm 0.82$ & \bm{$8.33 \pm 1.25$} & $0.67 \pm 0.47$ \\
    \midrule
    \multirow{7}{*}{\centering Max Error}
      & \textsc{MF-BoN+Random}
      & $1.43 \pm 0.16$ & $2.04 \pm 0.04$ & $2.57 \pm 0.04$
      && $1.74 \pm 0.06$ & $2.37 \pm 0.08$ & $2.82 \pm 0.04$ \\
    & \textsc{BoN+Random}
      & $1.31 \pm 0.10$ & $1.95 \pm 0.05$ & $2.48 \pm 0.04$
      && $1.69 \pm 0.05$ & $2.17 \pm 0.04$ & $2.64 \pm 0.02$ \\
    & \textsc{BS+Random}
      & $1.29 \pm 0.14$ & $1.90 \pm 0.05$ & $2.46 \pm 0.06$
      && $1.54 \pm 0.08$ & $2.19 \pm 0.06$ & $2.61 \pm 0.03$ \\
    & \textsc{GS+Random}
      & $1.27 \pm 0.14$ & $1.89 \pm 0.04$ & $2.41 \pm 0.01$
      && \bm{$1.45 \pm 0.11$} & $2.11 \pm 0.03$ & $2.58 \pm 0.08$ \\
    [-2pt] \cmidrule(lr){2-9}
    & \textsc{BoN+H} (ours)
      & $1.24 \pm 0.11$ & $1.87 \pm 0.06$ & $2.44 \pm 0.08$
      && $1.54 \pm 0.07$ & $2.14 \pm 0.07$ & $2.57 \pm 0.02$ \\
    & \textsc{BS+H} (ours)
      & \bm{$1.13 \pm 0.08$} & $1.90 \pm 0.05$ & $2.40 \pm 0.10$
      && $1.52 \pm 0.04$ & \bm{$1.99 \pm 0.07$} & $2.60 \pm 0.05$ \\
    & \textsc{GS+H} (ours)
      & $1.15 \pm 0.06$ & \bm{$1.85 \pm 0.05$} & \bm{$2.37 \pm 0.10$}
      && $1.49 \pm 0.05$ & $2.03 \pm 0.03$ & \bm{$2.50 \pm 0.05$} \\
    \midrule
    \multirow{7}{*}{\centering L2 Dist.}
      & \textsc{MF-BoN+Random}
      & $1.43 \pm 0.16$ & $2.30 \pm 0.05$ & $3.10 \pm 0.06$
      && $1.74 \pm 0.06$ & $2.71 \pm 0.09$ & $3.44 \pm 0.05$ \\
    & \textsc{BoN+Random}
      & $1.31 \pm 0.10$ & $2.18 \pm 0.08$ & $2.95 \pm 0.05$
      && $1.69 \pm 0.05$ & $2.48 \pm 0.05$ & $3.27 \pm 0.02$ \\
    & \textsc{BS+Random}
      & $1.29 \pm 0.14$ & $2.14 \pm 0.08$ & $2.94 \pm 0.08$
      && $1.54 \pm 0.08$ & $2.51 \pm 0.05$ & $3.25 \pm 0.08$ \\
    & \textsc{GS+Random}
      & $1.27 \pm 0.14$ & $2.12 \pm 0.08$ & $2.89 \pm 0.01$
      && \bm{$1.45 \pm 0.11$} & $2.41 \pm 0.02$ & $3.22 \pm 0.08$ \\
      [-2pt] \cmidrule(lr){2-9}
    & \textsc{BoN+H} (ours)
      & $1.24 \pm 0.11$ & $2.08 \pm 0.07$ & $2.88 \pm 0.11$
      && $1.54 \pm 0.07$ & $2.44 \pm 0.07$ & $3.20 \pm 0.03$ \\
    & \textsc{BS+H} (ours)
      & \bm{$1.13 \pm 0.08$} & $2.09 \pm 0.07$ & $2.84 \pm 0.11$
      && $1.52 \pm 0.04$ & \bm{$2.25 \pm 0.07$} & $3.17 \pm 0.07$ \\
    & \textsc{GS+H} (ours)
      & $1.15 \pm 0.06$ & \bm{$2.06 \pm 0.09$} & \bm{$2.82 \pm 0.14$}
      && $1.49 \pm 0.05$ & $2.29 \pm 0.04$ & \bm{$3.09 \pm 0.06$} \\
    \bottomrule
  \end{tabular}
  \end{adjustbox}
  \caption{Alignment performance results for $N=128$ prompt optimizer queries and $K=1$ response model query on different goal objective sizes for the reward model domains, \textsc{HelpSteer} and \textsc{Political}. Performance was evaluated on goal completion, max-error, and Euclidean distance reward functions. The \texttt{Llama-3.3-70B-Instruct} \cite{grattafiori2024llama3herdmodels} model was used as the prompt optimizer and the response model.}
  \label{table:large-k-1}
\end{table*}

\subsection{Baselines}
We compare \textsc{HIA} against three tuning‑free, inference‑time alignment strategies that vary in search methodology, reward signal, and prompt optimizer policy. All baselines are executed under the same inference budget as \textsc{HIA} (i.e.\ an identical number of calls to the prompt optimizer and to the response model) to ensure a fair cost–performance comparison.

\textsc{BoN} - A best-of-$N$ sampler that draws $N$ modified prompt variants from the prompt optimizer and selects the top prompt according to reward model evaluations. This baseline isolates the benefit of heuristic pre-filtering used by \textsc{HIA}.
    
\textsc{MF-BoN} - A Modification-Free (MF) version of \textsc{BoN} that skips the prompt optimizer and requests $K$ response completions using the initialized prompt. The original prompt and top response evaluated with the reward model is used for evaluation. This baseline evaluates the effectiveness of using sampling without a prompt optimizer policy.

\textsc{BS} - Standard beam search (BS) with \textit{beam width} $W$, \textit{depth} $D$, and \textit{branching factor} $N$. In this version of \textsc{BS}, we follow by the same methodology in \textsc{BoN} at each step, generating $N$ candidates, and selecting the top-$W$ modified prompt variants for the next step. For each beam, we append the selected top candidate and its reward to its history and append the history to the current state, allowing the prompt optimizer policy to adapt based on previous attempts. This baseline provides a multi-step history-dependent search strategy that represents works that incorporate beam-search for inference-time alignment \cite{singla2024dynamic}

\textsc{GS} - Greedy search (GS) with \textit{depth} $D$, and \textit{branching factor} $N$. It is a special case of \textsc{BS} with $W=1$ and extended depth $D$ under equivalent budgets.

For each baseline method, we apply both a \textsc{Random} and \textsc{H} (Heuristic) filtering policy. The \textsc{Random} filtering is meant to simulate existing sampling approaches that do not employ a heuristic filtering stage.

\subsection{Experimentation Setup}
For all experiments, the \texttt{Llama-3.3-70B-Instruct} \cite{grattafiori2024llama3herdmodels} local model as both the response model $\pi_\theta$ and the prompt optimizer $\pi_\phi$ (see Appendix for prompt optimizer ablation tests). We ran inference for the response model, prompt optimizer, and reward model using a \texttt{vLLM} server \cite{kwon2023efficient} on 16 A100 GPUs\footnote{Our experiments can be replicated with as little as 2 A100s}.

Each run, for a specific number of objectives and method, was generated on three static seeds each containing a set of 100 prompts to ensure reproducibility, and we report the mean and standard deviation of each run. For goal-conditioning, we uniformly sample random objectives and take a uniform random sample for each type of objective from the reward models' discrete scoring range. To facilitate fair comparisons between baselines, we configured each baseline's hyperparameters to match inference budgets for the prompt optimizer and the response model. Note that we define the cost of reward model inference as negligible due to their small size and therefore do not include it in the budget. For prompt optimizer and response model generation, we cap the maximum prompt length at 256 and the response length at 512 tokens.

\subsection{Sample Efficiency of \textsc{HIA}}
We analyze the sample efficiency of \textsc{HIA} on BoN sampling, beam search, and greedy search with respect to the number of queries made to the response model, showing that \textsc{HIA} boosts the sample efficiency of existing search and sampling methods especially on small budgets for response model inference (e.g., $K=1,2,4$). To establish fair comparisons between baselines, we set a budget of $N=128$ prompt optimizer completions and up to $K=16$ response model calls, leaving the $K$-selection to the filtering process--\textsc{Random} or \textsc{H}. For \textsc{BS}, we set a depth $D=2$, beam width $W=2$, branching factor $N'=32$, and $K'=4$ response inference calls per step. Similarly, in \textsc{GS}, we set a depth $D=4$, beam width $W=1$, branching factor $N'=32$, and $K'=4$.

In Table \ref{table:large-k-1}, we see a general increase in goal completion performance among BoN, beam search, and greedy search when a heuristic filter $H$ is applied. We find this trend to be common across both \textsc{HelpSteer} and \textsc{Political} on goal completion, Euclidean-distance, and max-error as well as across varying numbers of objectives when $K=1$, indicating definitive alignment improvement. For example, we see a $29\% \,(24 \rightarrow 31)$ \textsc{HelpSteer} goal completion improvement on single-objective optimization for BoN sampling, $18.4\% \,(25.33 \rightarrow 30)$ for beam search, and $12.4\% \, (26.67 \rightarrow 30)$ for greedy search.

We also find that beam search and greedy search tend to outperform BoN regardless of the filtering policy employed under equivalent inference budgets, confirming that search methodologies are preferable in practice due to their in-context learning and exploration ability.

\begin{figure*}[t]
    \centering
    \includegraphics[width=\linewidth]{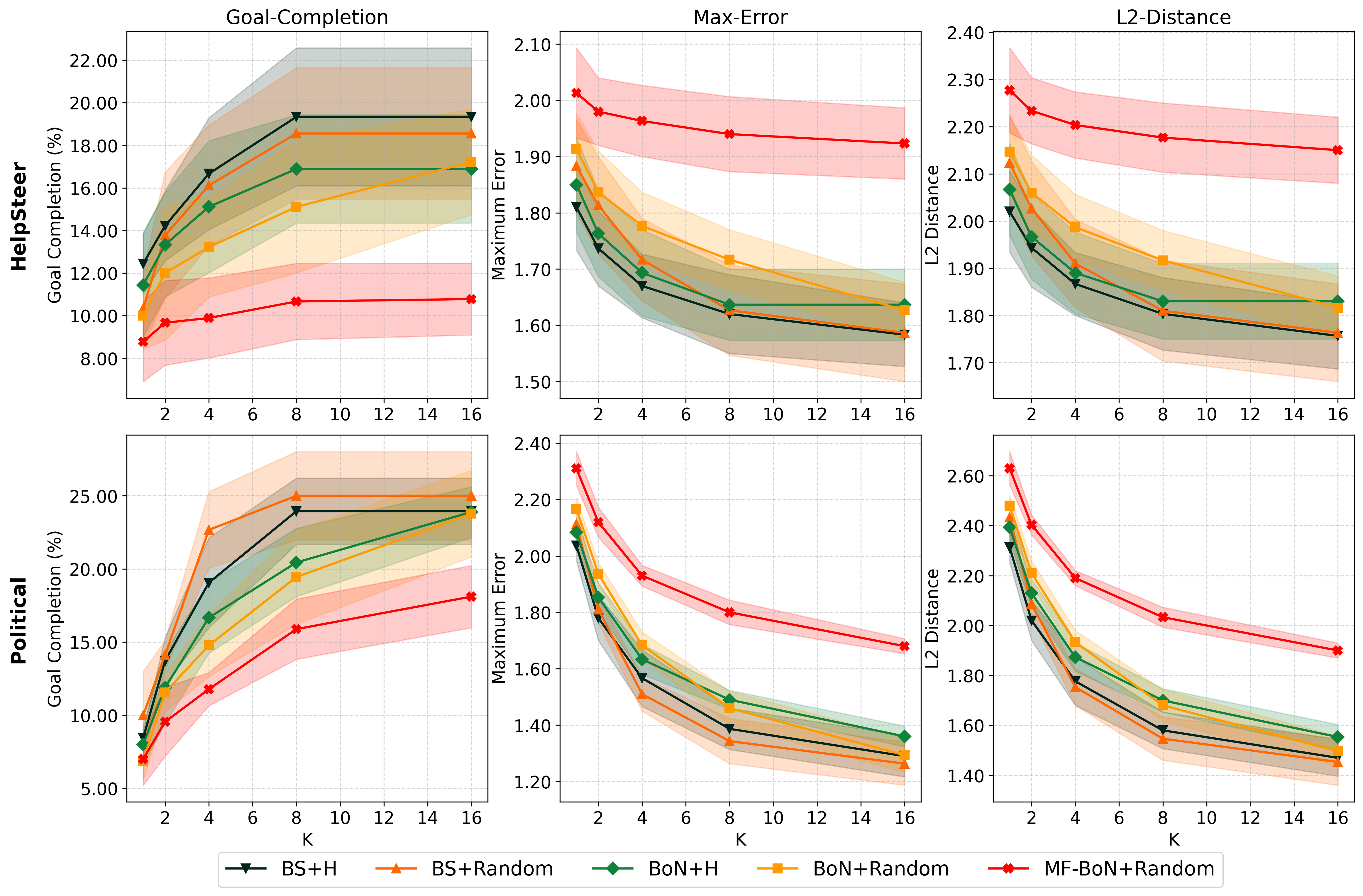}
    \caption{Alignment performance over $K$ number of response model queries using $N=128$ prompt optimizer queries on the \textsc{HelpSteer} and \textsc{Political} domains and evaluated on three reward functions. Each subplot is averaged over three goal objective sizes $(1, 2, 3)$. The \texttt{Llama-3.3-70B-Instruct} \cite{grattafiori2024llama3herdmodels} model was used as the prompt optimizer and the response model.}
    \label{fig:combined_domains_triptych}
\end{figure*}

\subsection{\textsc{HIA} for Large Inference Budgets}
Although \textsc{HIA} improves sample efficiency when $K$ is small, this is not generally the case when $K$ is large (e.g., $K=8,16$). We observe in Figure \ref{fig:combined_domains_triptych} that \textsc{Random} filtering approaches begin to outperform heuristic filtering as $K$ approaches $16$. This phenomenon is especially present in BoN methods in both \textsc{HelpSteer} and \textsc{Political} and across all three reward functions where \textsc{BoN+H} exhibits better alignment performance over \textsc{BoN+Random} until a threshold $K$ value. We hypothesize that due to the approximation gap between the heuristic reward models and the reference reward models, increasing $K$ will result in performance degradation for \textsc{H} baselines due to greater sample diversity in the \textsc{Random} baselines.

\section{Discussion}
In this section, we discuss the assumptions, limitations, and potential improvements of \textsc{HIA}.

\textbf{Assumptions and Limitations.}
Throughout this paper, we assume all rewards hold descriptive, interpretable meaning with respect to their reward models. This may not necessarily hold true for reward models trained on context-independent preferences that tend to be ambiguous. Prompt optimization, as applied in \textsc{HIA}, requires detailed reward context to avoid spurious prompt modifications that can bottleneck alignment performance.

To simplify our experiments, we assume the response model fully adheres to the prompt without considering response refusal due to potential harm or safety violations. We chose \texttt{Llama-3.3-70B-Instruct} as the response model since \texttt{Llama} models had the lowest refusal rates among local models on the \textsc{Sorry-Bench} \cite{xie2024sorry}. This is not an impractical assumption for black-box models since the models with the least refusal rates were black-box on the \textsc{Sorry-Bench}. 

\textbf{Potential Improvements.}
A simple approach to improving \textsc{HIA} performance is additional fine-tuning of the heuristic reward models. We trained on approximately 100k samples for \textsc{HelpSteer} and 200k for \textsc{Political} for each objective, which we showed to be effective for small inference budgets. However, improving performance for larger inference budgets may rely on reducing the approximation gap between reference and heuristic reward models, requiring more fine-tuning on datasets with higher quality trajectories.

To improve generalization across many response models, the heuristic reward model training datasets will require samples from a diverse set of response models which can become costly if generalization is prioritized. However, if the target response models are known prior, then trajectory generation costs can be mitigated.

Additionally, \textsc{HIA} can incorporate nonstationary reward models \cite{mahmud2025maple} that can be actively learned at inference-time, pairing well with our fine-tuning-free framework and facilitating better personalized alignment on out-of-distribution users.

\section{Conclusion}
We introduced \textsc{HIA}, a search-agnostic and tuning-free framework for inference-time alignment that pairs a prompt optimizer with low-cost heuristic reward models for filtering. Under equal compute budgets, \textsc{HIA} maintains competitive performance with as few as one or two response-policy queries and delivers consistent improvement across goal completion, max-error, and Euclidean distance reward functions for up to three simultaneous objectives. Our results on two real-world prompt datasets, \textsc{HelpSteer} and \textsc{ComPRed} indicate that employing a heuristic filtering step can provide improved sample efficiency for low-inference budgets, an essential property for practical, personalized inference-time alignment at deployment time.

\section*{Acknowledgments}
This research was supported in part by the U.S.~Army DEVCOM Analysis Center (DAC) under contract number W911QX23D0009, and by the National Science Foundation
under grants 2205153, 2321786, and 2416460.

\bibliography{aaai2026}

\section{Appendix A -- Additional Figures}

\subsection{Prompt Optimizer Model Ablation}
\begin{figure}[!h]
    \centering
    \includegraphics[width=\linewidth]{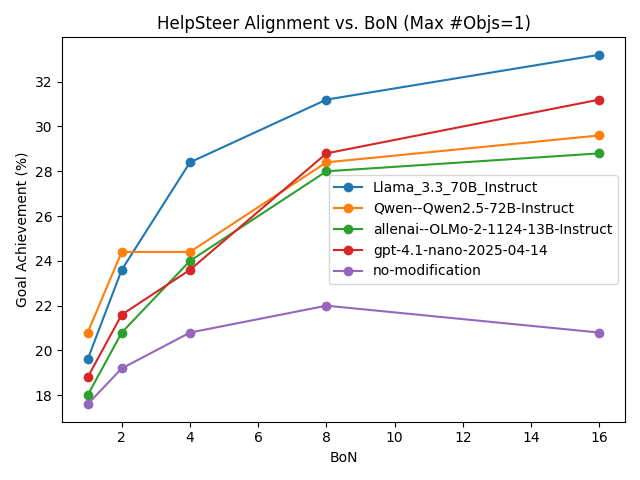}
    \caption{An ablation on different prompt modifier models for varying numbers of BoN samples, using \texttt{Llama-3.3-70B-Instruct} as the response model.}
    \label{fig:enter-label}
\end{figure}

\subsection{Performance Tradeoff of Heuristic Reward Models} 

\begin{figure}[!h]
    \centering
    \includegraphics[width=\linewidth]{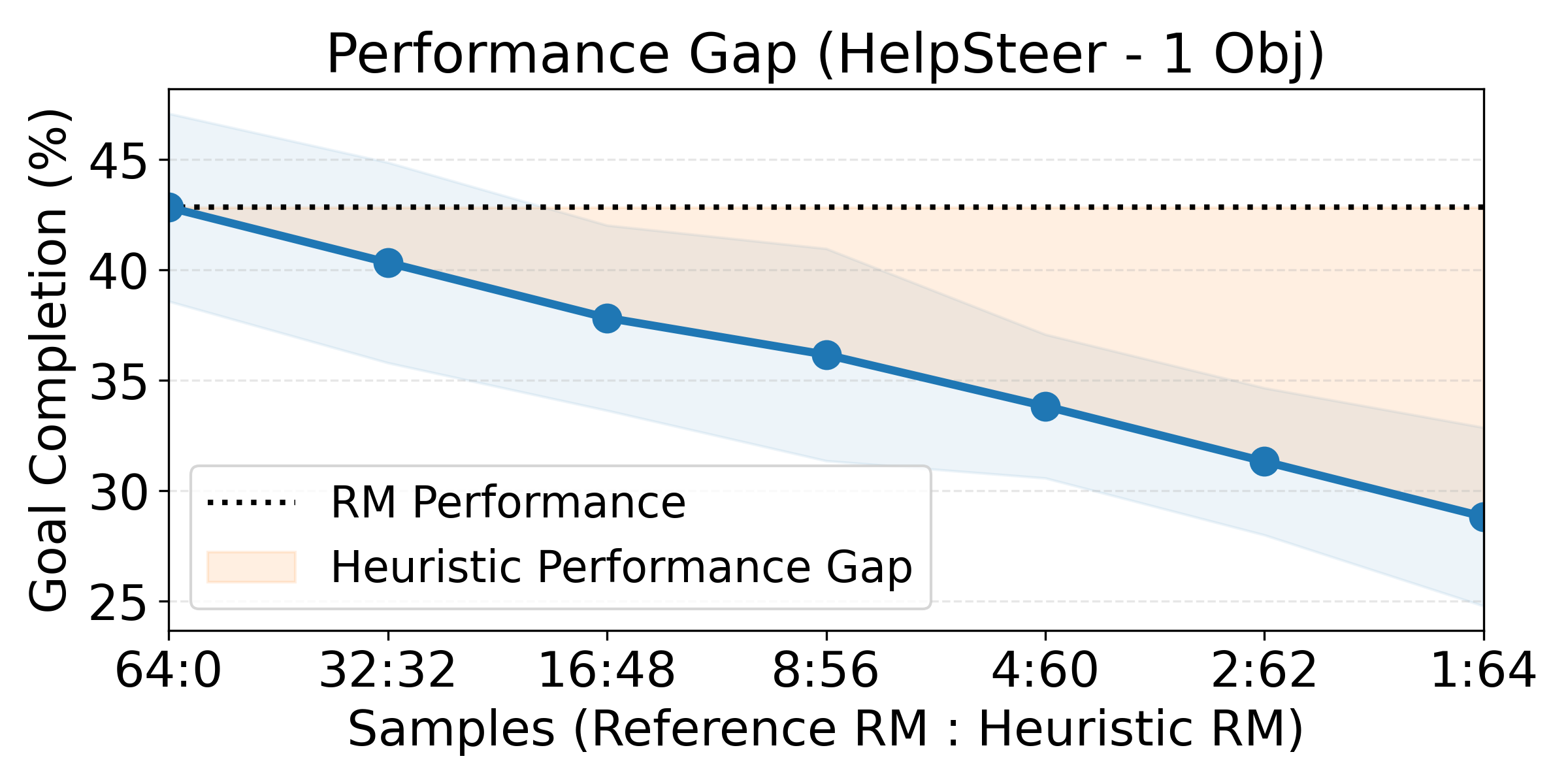}
    \caption{BoN goal completion rate when swapping number of samples between the heuristic reward model and the reference reward model.}
    \label{fig:efficiency_helpsteer_v2}
\end{figure}

\begin{figure*}[!h]
    \centering
    \includegraphics[width=\linewidth]{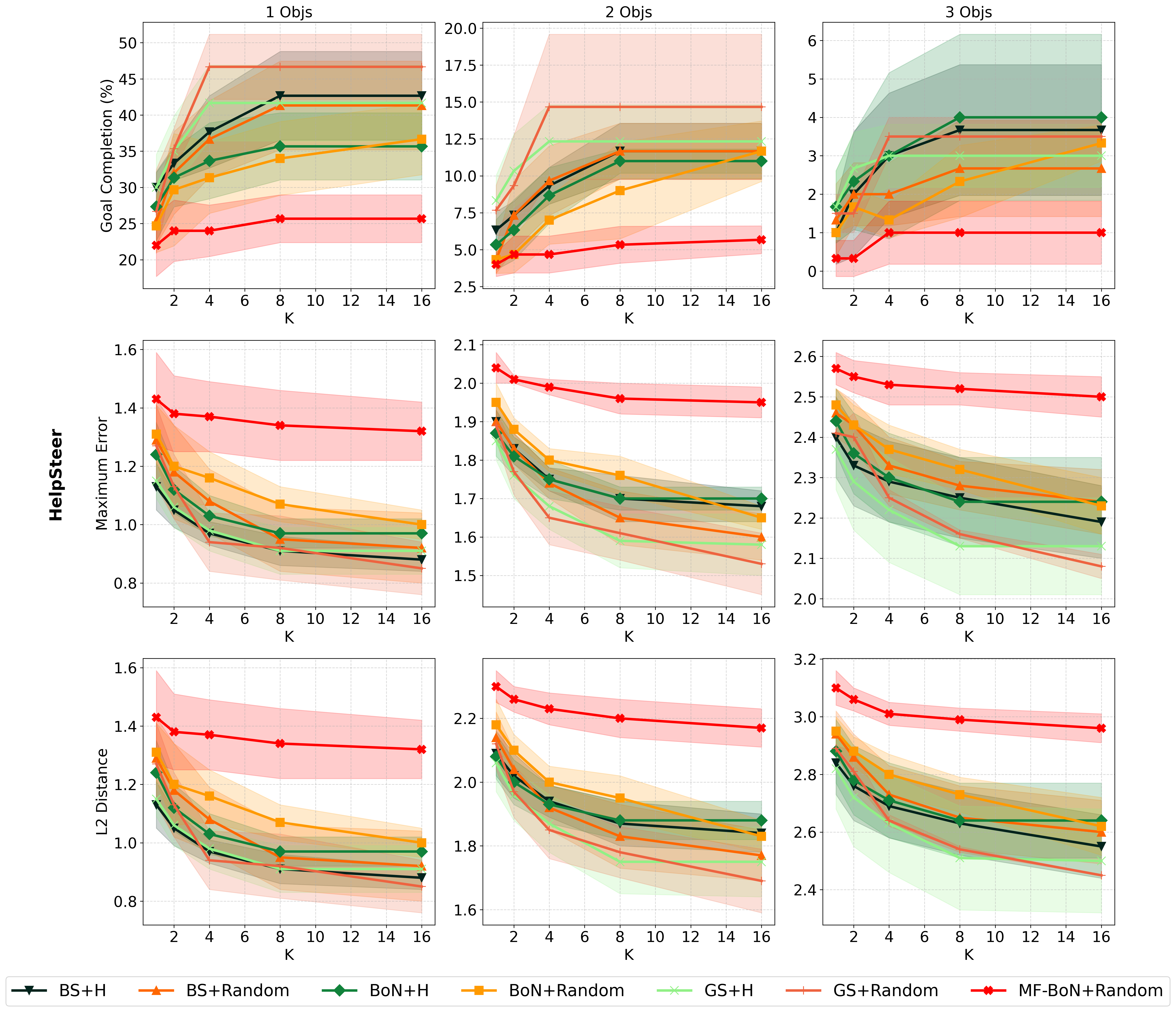}
    \caption{Alignment performance over $K$ number of response model queries using $N=128$ prompt optimizer queries on the \textsc{HelpSteer} domain and evaluated on three reward functions over three goal objective sizes. The \texttt{Llama-3.3-70B-Instruct} model was used as the prompt optimizer and the response model.}
    \label{fig:3x3_helpsteer}
\end{figure*}

\begin{figure*}[!h]
    \centering
    \includegraphics[width=\linewidth]{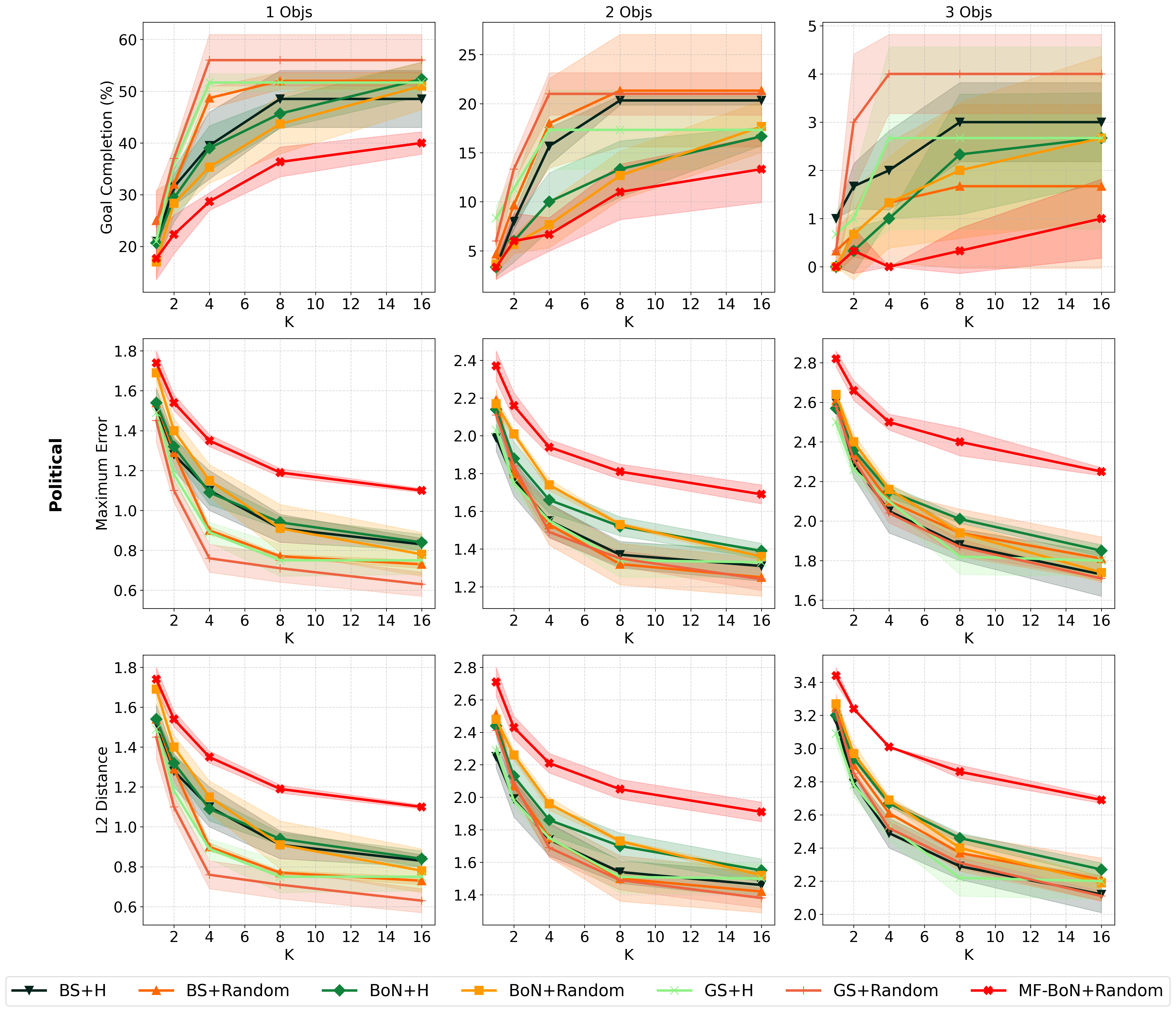}
    \caption{Alignment performance over $K$ number of response model queries using $N=128$ prompt optimizer queries on the \textsc{Political} domain and evaluated on three reward functions over three goal objective sizes. The \texttt{Llama-3.3-70B-Instruct} model was used as the prompt optimizer and the response model.}
    \label{fig:3x3_political}
\end{figure*}

\section{Appendix B -- Experiment Details}

\subsection{Hyperparameters}
\begin{table}[!h]
  \centering
  \setlength\tabcolsep{5pt}
\renewcommand{\arraystretch}{1.0}
\resizebox{\columnwidth}{!} {
\begin{tabular}{@{}cccccc@{}}
    \toprule
    Method & $N$ & $K$ & $W$ & $D$ & Seeds\\
    \midrule
      \textsc{MF-BoN+[Random, H]}  & 128 & 16 & -- & -- & [888, 89, 12] \\
      \textsc{BoN+[Random, H]}  & 128 & 16 & -- & -- & [888, 89, 12] \\
      \textsc{BS+[Random, H]}  & 128 & 16 & 2 & 2 & [888, 89, 12] \\
      \textsc{GS+[Random, H]}  & 128 & 16 & 1 & 4 & [888, 89, 12] \\
    \bottomrule
\end{tabular}}
\caption{Hyperparameter details.}
\end{table}

\begin{table*}[h]
  \centering
  \renewcommand{\arraystretch}{1.15}
  \setlength{\tabcolsep}{6pt}
  \begin{tabularx}{\textwidth}{>{\raggedright\arraybackslash}p{3.8cm} X}
    \toprule
    \textbf{Field} & \textbf{Content} \\
    \midrule
    Original Prompt &
    I own a medical spa. I would like to set up protocols for services I perform with instructions for me on how to use each device. Please give me step-by-step instructions on how to use the Ensculpt Neo on a client. \\[10pt] \midrule

    Best Modified Prompt &
    Given the context of being an expert in medical aesthetics and having access to an Ensculpt Neo device, please provide step-by-step, clear, and logically organized instructions on how to use the Ensculpt Neo on a client, ensuring that each step is concise, easy to understand, and seamlessly connected to achieve a comprehensive and coherent guide for the procedure. \\[10pt] \midrule

    Top 5 Candidate Prompts &
      \textbf{1.}~Given the context of being an expert in medical aesthetics and having access to an Ensculpt Neo device, please provide step-by-step, clear, and logically organized instructions on how to use the Ensculpt Neo on a client, ensuring that each step is concise, easy to understand, and seamlessly connected to achieve a comprehensive and coherent guide for the procedure.\par
      \textbf{2.}~I have a company and I need clear step-by-step instructions. I own a medical spa. I would like to set up protocols for services I perform with instructions for me on how to use each device. Please provide me with a highly organized, clear, and easy-to-follow step-by-step guide on how to use the Ensculpt Neo on a client, ensuring that every detail is well-explained and easy to understand.\par
      \textbf{3.}~Please provide a detailed, thoroughly structured, step-by-step guide on how to use the Ensculpt Neo device for a client at my medical spa, including clear and concise language, numbered or bullet-pointed instructions where applicable, and ensuring that the information is presented in a logical, easy-to-follow order to minimize confusion and maximize understanding.\par
      \textbf{4.}~To effectively utilize the Ensculpt Neo for the benefit of your medical spa, I would appreciate it if you could provide me with a clear, step-by-step guide on how to use the Ensculpt Neo on a client, including any necessary preparation, the treatment process itself, and post-treatment protocols, all presented in a highly organized and easy-to-follow manner to ensure I can understand and implement the instructions correctly. \par
      \textbf{5.}~What are the detailed steps to set up the company and provide a step-by-step guide on how to use the Ensculpt Neo on a client in my medical spa with clear, organized, and easy-to-follow instructions to ensure I can easily understand and implement the protocol? \\[10pt] \midrule

    Optimization Goal & \textbf{coherence} (weight = 4) \\[2pt] \midrule
    Best Achieved Reward & \textbf{coherence} (weight = 4) \\[2pt] \midrule

    Best Response &
    \textbf{Ensculpt Neo Treatment Protocol: A Step-by-Step Guide}\par\smallskip
    \textbf{Pre-Treatment Preparation (Steps 1–5)}\par
    1.~\textit{Client Consultation}: Conduct a thorough consultation to discuss the client’s concerns, medical history, and treatment goals. Ensure they understand the Ensculpt Neo procedure, benefits, and potential side effects.\par
    2.~\textit{Informed Consent}: Obtain signed informed consent, covering treatment details, potential risks, and after-care instructions.\par
    3.~\textit{Prepare the Treatment Area}: Clean and disinfect the target area; remove any lotions, oils, or makeup.\par
    4.~\textit{Client Positioning}: Position the client comfortably so the treatment area is fully accessible.\par
    5.~\textit{Device Preparation}: Power on the Ensculpt Neo, perform a system check, and calibrate per manufacturer guidelines.\par\medskip

    \textbf{Treatment Setup and Administration (Steps 6–10)}\par
    6.~\textit{Applicator Selection}: Choose the correct applicator size/type for the area (abdomen, thigh, arm, etc.).\par
    7.~\textit{Applicator Placement}: Align and secure the applicator ensuring full-skin contact.\par
    8.~\textit{Parameter Settings}: Program energy level, pulse duration, and frequency according to skin type, area, and desired outcome.\par
    9.~\textit{Initiate Treatment}: Start the session; the device delivers electromagnetic energy to induce muscle contractions and fat reduction.\par
    10.~\textit{Monitor Client}: Continually assess client comfort and tissue response, adjusting parameters as needed... \\
    \bottomrule
  \end{tabularx}
  \caption{Prompt optimization example for \textsc{HelpSteer}}
  \label{tab:ensculpt-neo-row8}
\end{table*}

\begin{table*}[!h]
  \centering
  \setlength\tabcolsep{5pt}
\renewcommand{\arraystretch}{1.0}
\resizebox{\textwidth}{!} {
\begin{tabular}{@{}ccccc@{}}
    \toprule
    Policy & HF ID (base model) & Temperature & Max Tokens & Parameter Count\\
    \midrule
    Prompt Optimizer & \texttt{meta-llama/Llama-3.3-70B-Instruct} & 0.7 & 256 & 70.6B \\
    Response & \texttt{meta-llama/Llama-3.3-70B-Instruct} & 0.7 & 512 & 70.6B\\
    Heuristic Reward Model  & \texttt{answerdotai/ModernBERT-large} & -- & -- & 396M \\
    Reference Reward Model  & \texttt{answerdotai/ModernBERT-large} & -- & -- & 396M \\
    \bottomrule
\end{tabular}}
\caption{Model/Policy details.}
\end{table*}

\begin{table*}[!h]
  \centering
  \setlength\tabcolsep{5pt}
\renewcommand{\arraystretch}{1.0}
\resizebox{\textwidth}{!} {
\begin{tabularx}{\textwidth}{@{}cccX@{}}
    \toprule
    Dataset & Reward Model Name & Discrete Reward Range & Prompt Description\\
    \midrule
      \multirow{19}{*}{\centering \textsc{HelpSteer}}  & Verbosity &  [0,4] & The level of detail and elaboration in the response relative to the prompt's requirements. A score of 0 encourages brief, **minimal** responses that provide only the bare essentials or less. A score of 4 demands highly detailed, **expansive** responses that include extensive elaboration, examples, or supplementary information beyond the minimum needed to address the prompt. \\ [-2pt]\cmidrule(lr){2-4}
          & Coherence & [0,4] &  Consistency and clarity of expression. A score of 0 implies an **incoherent**, not logical, or not understandable responses (e.g, gibberish). A score of 4 implies a highly organized, seamless, and clear response that can be **easily** be interpreted by the user. \\ [-2pt]\cmidrule(lr){2-4}
          & Complexity & [0,4] &  The intellectual depth and specialized knowledge of the response, reflecting the level of expertise needed. A score of 0 allows for **simplistic**, general-knowledge responses that anyone with basic language skills could produce. A score of 4 requires responses that demand **advanced** domain expertise, incorporating technical, nuanced, or sophisticated concepts specific to the subject matter. \\ \midrule
      \multirow{20}{*}{\centering \textsc{Political}}  & Left & [0,5] & Measures how much the response is aligned with Left political ideologies. A score of 0 implies a response that is **not** politically aligned with Left ideology and is instead **apolitical**. A score of 3 implies a response that is **slighlty** politically aligned with Left ideology. A score of 5 implies a response is **very** politically aligned with Left ideology. \\ [-2pt]\cmidrule(lr){2-4}
          & Center & [0,5] &  Measures how much the response is aligned with Center political ideologies. A score of 0 implies a response that is **not** politically aligned with Center ideology and is instead **apolitical**. A score of 3 implies a response that is **slighlty** politically aligned with Center ideology. A score of 5 implies a response that is **very** politically aligned with Center ideology. \\ [-2pt]\cmidrule(lr){2-4}
          & Right & [0,5] &  Measures how much the response is aligned with Right political ideologies. A score of 0 implies a response that is **not** politically aligned with Right ideology and is instead **apolitical**. A score of 3 implies a response that is **slighlty** politically aligned with Right ideology. A score of 5 implies a response that is **very** politically aligned with Right ideology. \\
    \bottomrule
\end{tabularx}}
\caption{Reward model details.}
\end{table*}

\begin{table*}[!h]
  \centering
  \setlength\tabcolsep{5pt}
\renewcommand{\arraystretch}{1.0}
\resizebox{\textwidth}{!} {
\begin{tabularx}{\textwidth}{@{}cX@{}}
    \toprule
    Prompt Type & Prompt\\
    \midrule
      \multirow{45}{*}{\centering \textsc{System}}  &  Ignore all previous instructions. You are a prompt-modifier agent. Your task is to refine the user's prompt based on the provided goals and constraints.\newline
        \# Steps \newline
        1. Understand the Task: Grasp the main objective, goals, requirements, constraints, and expected output.\newline
        2. Task: Modify the orginal prompt to elecit a response that achieves the goal objective weights.\newline
        3. Context: For each objective weight, a score of 0 indicates absolutley no presence of the objective in the response while a score of 5 indicates maximal presence of the objective in the response.\newline
        \# If using beam search or greedy search; otherwise, omit \newline
        4. Utilizing the History: If non-empty, use the historical samples of modified prompts and their associated scores to improve your current modified prompt.\newline
        
        Constraints:\newline
        - Output **only** the block with your modified prompt inside with no commentary before or after.:\newline
          $<$modified\_prompt$>$\verb|\n| ... \verb|\n|$<$/modified\_prompt$>$\newline
        - The block **must contain a complete modified prompt that does not deviate from the intent of the original prompt**.\newline
        - If the original prompt is a question, make the modified prompt a question as well.\newline
        - Do **not** mention these meta-instructions or goal objective weights.\newline
        
        Example:\newline
        $<$original\_prompt$>$\newline
        How do I start my own company?\newline
        $<$/original\_prompt$>$\newline
        
        $<$goal\_objective\_weights$>$\newline
        (verbosity: 5, right: 1)\newline
        $<$/goal\_objective\_weights$>$\newline
        
        $<$descriptor\_details$>$\newline
        Verbosity: The level of detail and elaboration in the response relative to the prompt's requirements. A score of 0 encourages brief, minimal responses that provide only the bare essentials or less. A score of 2 offers moderate amounts of detail and explanation. A score of 4 demands highly detailed, expansive responses that include extensive elaboration, examples, or supplementary information beyond the minimum needed to address the prompt.\newline
        Right: Measures how much the response is aligned with Right political ideologies. A score of 0 implies a response that is **not** politically aligned with Right ideology and is instead **apolitical**. A score of 3 implies a response that is **slighlty** politically aligned with Right ideology. A score of 5 implies a response that is **very** politically aligned with Right ideology.\newline
        $<$/descriptor\_details$>$\newline
        
        $<$modified\_prompt$>$\newline
        How do I start my own company? Give me a highly detailed explanation while offering examples of companies that align with the right political ideology.
        $<$/modified\_prompt$>$\newline
        Format the output as:\newline
        $<$modified\_prompt$>$\verb|\n| [modified\_prompt] \verb|\n|$<$/modified\_prompt$>$ \\ \midrule
      \multirow{15}{*}{\centering \textsc{User}}  & Given the following details, generate a modified prompt.\newline
        $<$original\_prompt$>$\newline
        \{original\_prompt\}\newline
        $<$/original\_prompt$>$\newline
        
        $<$goal\_objective\_weights$>$\newline
        \{goal\}\newline
        $<$/goal\_objective\_weights$>$\newline
        
        $<$descriptor\_details$>$\newline
        \{descriptor\_str\}\newline
        $<$/descriptor\_details$>$\newline

        \# If using beam search or greedy search; otherwise, omit\newline
        $<$history$>$\newline
        \{history\}\newline
        $<$/history$>$ \\
    \bottomrule
\end{tabularx}}
\caption{System and user prompts.}
\end{table*}


\begin{figure*}[!h]                  
  \centering
  \small
  \begin{lstlisting}[language=Python,
                     caption={Simplified Best-of-$N$ (BoN) sampling with HIA pseudo-code},
                     label={lst:bon}]
"""
Simplified Best-of-N (BoN) sampling loop with HIA.
For each user prompt:
    1. Generate N candidate modified prompts from a prompt optimizer model.
    2. Score each modified prompt candidate with the heuristic reward model(s) response ID + modified prompt
    3. Filter the top-K candidates
    4. Send top-K candidates through response model
    5. Score each K-candidate with the reference reward model(s) using response + modified prompt
    6. Keep the single best-scored modified prompt + response.
"""

import random
from typing import List, Tuple

PROMPT_OPTIMIZER = model(...)
RESPONSE_MODEL = model(...)
HRM = model(...)
RRM = model(...)
def generate_candidates(prompt: str, n: int, prompt_optimizer: Model) -> List[str]:
    """Generate modified prompt candidates"""
    candidates = prompt_optimizer(prompt, n_completions=n)
    return candidates

def generate_resopnse(prompt: str, response_model: Model) -> str
    return response_model(prompt)
    
def reference_reward_model(prompt: str, response: str, rrm: Model) -> float:
    """Score <prompt, response> with a reference reward model."""
    return rrm(prompt, response)
    
def heuristic_reward_model(prompt: str, response_id: str, hrm: Model) -> float:
    """Score <prompt, response_id> with a heuristic reward model."""
    return hrm(prompt, response_id)

def best_of_n_with_H(prompt: str, n: int, k: int) -> Tuple[str, float]:
    """Return the top-scoring response (and its score) among N candidates."""
    # Get modified prompt candidates
    candidates = generate_candidates(prompt, n, PROMPT_OPTIMIZER)
    # Score top-K candidates
    h_scored = [(resp, heuristic_reward_model(modified_prompt, response_model_id, HRM)) for modified_prompt in candidates]
    # Get top-K modified prompts
    sorted_h_scored = sort(h_scored, key=lambda t: t[1])
    top_k = sorted_h_scored[:k]
    # Get responses for top-K candidate prompts
    responses = [generate_resopnse(modified_prompt, RESPONSE_MODEL) for modified_prompt in top_k]
    # Score each response with the reference reward model
    ref_scored = [(response, reference_reward_model(prompt, response, RRM)) for response in responses] 
    sorted_ref_scored = sort(ref_scored, key=lambda t: t[1])
    # Return the response with the best reference score
    best_response, best_ref_score = sorted_ref_scored[:1]
    return best_response, best_ref_score

if __name__ == "__main__":
    random.seed(888)
    # Number of modified prompts
    N = 128  
    # Number of modified prompt candidates to choose
    K = 16
    prompts = [
        "Summarise the plot of Dune in one paragraph.",
        "List three benefits of functional programming.",
        "Translate 'good morning' into Japanese.",
    ]
    for p in prompts:
        best_response, score = best_of_n_with_H(p, N, K)
  \end{lstlisting}
  \vspace{-0.5em}
\end{figure*}

\end{document}